\documentclass[a4paper,twoside,french]{article}
\usepackage{rfia2000}
\usepackage[T1]{fontenc}
\usepackage{babel}
\usepackage{times}
\usepackage{graphicx}
\usepackage{amsmath}

\begin{document}
\date{}
\title{\Large\bf Suivi d'objets basé forme et couleur pour la navigation robotique en temps réel
       }
\author{Haythem Ghazouani \\
  LIRMM, Département de robotique \\
  Université de Montpellier II , 161 rue Ada, 34392 Montpellier Cedex 5, France \\
  haythemghz@yahoo.fr\\
	{\bf } \\
{\bf }}

\maketitle
\thispagestyle{empty}
\subsection*{R\'esum\'e}
{\em
Ce papier présente une approche temps-réel pour la détection et le suivi d'une balle unicolore. L'approche consiste en deux phases principales. Dans une première phase de calibrage, qui s'effectue hors ligne, les paramètres intrinsèques de la caméra ainsi que la distorsion radiale sont estimés, et une classification de couleurs est apprise à partir d'un exemplaire d'une image de balles colorées. La deuxième phase de suivi temps réel consiste en quatre étapes principales; (1) segmentation couleur de l'image d'entrée en plusieurs régions en se basant sur la classification hors-ligne, (2) estimation robuste des paramètres du cercle, (3) raffinement des paramètres du cercle, et (4) le suivi de la balle. Les résultats expérimentaux ont montré que l'approche présente un bon compromis entre adéquation à la navigation en temps réel et robustesse aux occultations, aux encombrements du fond ainsi qu'aux interférences de couleurs dans la scène.
}
\subsection*{Mots Clef}
Segmentation couleur, Détection d'objets, Suivi d'objets, Navigation visuelle.

\subsection*{Abstract}
{\em
This paper presents a real-time approach for single-colored ball detection and tracking. The approach consists of two main phases. In a first offline calibration phase, the intrinsic parameters of the camera and the radial distortion are estimated, and a classification of colors is learned from a sample image of colored balls. The second phase consists of four main steps: (1) color segmentation of the input image into several regions based on the offline classification, (2) robust estimation of the circle parameters (3) refinement of the circle parameters, and (4) ball tracking. The experimental results showed that the approach presents a good compromise between suitability for real-time navigation and robustness to occlusions, background congestion and colors interference in the scene.
}
\subsection*{Keywords}
Color segmentation, Object detection, Object tracking, Visual navigation.

\section{Introduction}

Le suivi d\textquoteright{}objets dans une séquence d\textquoteright{}images
occupe une place prépondérante dans plusieurs domaines en relation
avec la vision artificielle : surveillance, robotique, etc. Deux contraintes
se posent pour développer un algorithme de suivi robuste : la première
est la qualité du suivi et la deuxième est l\textquoteright{}aspect
temps réel exprimé via la rapidité et la complexité de l\textquoteright{}algorithme.
Nous nous intéressons dans ce papier au suivi d'objets basé sur la
couleur et la forme. Le suivi basé couleur a été principalement considéré
selon deux approches. Dans l\textquoteright{}algorithme du mean shift,
la recherche de l\textquoteright{}objet est effectuée en minimisant
une distance entre histogrammes de couleurs selon une méthode de type
descente de gradient, et conduit à une bonne précision de suivi. Cependant,
cette recherche étant déterministe, elle ne permet pas d\textquoteright{}être
robuste aux occultations importantes, et l\textquoteright{}algorithme
peut échouer en présence d\textquoteright{}un autre objet de couleurs
similaires, ou dans le cas de grands déplacements. Dans un autre cadre,
d'autres méthodes utilisent le même type de critère de ressemblance
entre histogrammes, mais l\textquoteright{}intègrent dans un filtre
particulaire. La densité de probabilité a posteriori de la position
de l\textquoteright{}objet est discrétisée en un ensemble de particules.
L\textquoteright{}évolution de ces particules et l\textquoteright{}estimation
de leur moyenne remplacent ici la minimisation effectuée dans l\textquoteright{}algorithme
du mean shift. L\textquoteright{}utilisation de ce cadre probabiliste
du filtrage particulaire induit une meilleure robustesse vis-à-vis
des occultations ou de la présence d\textquoteright{}objets similaires.
Cependant ces dernières méthodes restent difficiles à mettre au point
et peu adaptées au temps réel. 

Dans cet papier,  nous commençons par présenter une classification des méthodes de suivi d'objets. Ensuite, nous donnons une description détaillée de notre approche
de suivi d'objets et les expérimentations aux quelles elle a aboutit.

\section{Classification et critique des méthodes de suivi d\textquoteright{}objets}

Dans la littérature, de nombreuses méthodes de suivi d\textquoteright{}objets
ont été présentées ; une grande partie d\textquoteright{}entre elles,
peuvent être utilisées pour suivre des objets précis en temps réel
{[}1{]}.

Plusieurs classifications des méthodes de suivi visuel d\textquoteright{}objets
ont été proposées dans la littérature ; elles dépendent autant des
auteurs, que du but pour lequel ces méthodes ont été conçues. Nous
considérons la classification donnée dans {[}2{]}, où selon les
auteurs, les méthodes de suivi visuel peuvent être divisées en quatre
classes:
\begin{itemize}
\item \emph{Méthodes de suivi fondées sur des modèles}. Ces méthodes repèrent
des caractéristiques connues dans la scène et les utilisent pour mettre
à jour la position de l\textquoteright{}objet. Parmi ces méthodes,
citons celles qui exploitent les modèles géométriques fixes {[}3{]},
et les modèles déformables.
\item \emph{Méthodes de suivi de régions ou blobs.} Cette sorte de méthodes
se caractérise par la définition des objets d\textquoteright{}intérêt
comme ceux qui sont extraits de la scène en utilisant des méthodes
de segmentation. Citons les nombreuses méthodes qui détectent une
cible à partir de son mouvement sur un fond statique ou quasiment
statique {[}4{]}.
\item \emph{Méthodes de suivi à partir de mesures de vitesse.} Ces méthodes
peuvent suivre les objets en exploitant les mesures de leur vitesse
dans l\textquoteright{}image, avec des mesures telles que le flux
optique ou des équivalents {[}5{]}.
\item \emph{Méthodes de suivi de caractéristiques.} Ces méthodes suivent
certaines caractéristiques de l\textquoteright{}objet, comme des points,
des lignes, des contours\ldots{}{[}6{]}, caractéristiques ou primitives
image auxquelles il est possible aussi d\textquoteright{}imposer de
restrictions globales {[}7{]}. Ces caractéristiques peuvent être
aussi définies par la texture ou la couleur {[}8{]}.
\end{itemize}
Cette classification n\textquoteright{}est pas exhaustive, et à ce
jour, il existe de nombreux recouvrements entre les classes, c\textquoteright{}est-à-dire,
des méthodes qui peuvent être classifiées dans deux ou plusieurs classes.
Nous considérerons que ces méthodes sont des combinaisons des approches
existantes.

C\textquoteright{}est à partir des définitions des environnements
et des cibles pour chaque problématique de la navigation évoquée précédemment,
qu\textquoteright{}il est possible de s\textquoteright{}apercevoir,
que certaines méthodes, comme le suivi de blobs, seront difficilement
utilisables. Par ailleurs, il est très difficile d\textquoteright{}utiliser
des méthodes fondées sur la différence objet/fond, parce que le robot
est en mouvement et donc, le fond ou l\textquoteright{}arrière plan
n\textquoteright{}est pas statique (du point de vue de l\textquoteright{}image),
et même les cibles peuvent être statiques par rapport au fond.

De manière similaire, les méthodes fondées sur des mesures de vitesse,
seront difficilement exploitables pour la navigation de robots mobiles.
Il existe quelques approches pour le suivi à partir de flux optique,
qui ont été essayées pour la navigation d\textquoteright{}un robot
{[}9{]}; mais, la plupart d\textquoteright{}entre elles utilisent
une méthode de suivi de caractéristiques comme des lignes droites,
et c\textquoteright{}est à partir de ces primitives éparses dans l\textquoteright{}image,
que le calcul de la vitesse est fait.

Nous favorisons donc, l\textquoteright{}utilisation des deux autres
sortes de méthodes afin de réaliser les tâches de suivi depuis un
robot se déplaçant dans les environnements d'intérieur : suivi fondé
sur un modèle de la cible et suivi de primitives image. 
Nous proposons dans ce qui suit une approche robuste qui peut être
classée à la fois sous les deux classes retenues, à savoir le suivi
fondé sur un modèle de la cible et le suivi de primitives d'objets.
Notre approche se déroule en deux étapes; une première étape hors
ligne pour l'apprentissage du modèle d'objet (couleur et forme) et
une deuxième étape qui permet d'estimer les paramètres de l'objet
pour le suivi temps réel.

\section{Proposition d'une approche pour la détection et le suivi de balle
unicolore}

Dans cette section nous introduisons notre nouvelle approche pour
le suivi temps réel d'une balle colorée. L'idée clef qui permet de
trouver un compromis entre la robustesse et la rapidité du traitement
est la détection de la couleur basée sur une segmentation couleur
rapide qui produit un nombre beaucoup plus réduit de pixels de contours
par rapport à des approches standard basées sur la luminance{[}13, 14, 15{]}. Cette
réduction diminue considérablement le nombre de votes requis pour
une détection robuste en temps réel des paramètres du cercle, même
dans le cas où de nombreuses balles de couleur sont présentes dans
l'image.

L'approche consiste en deux phases principales. Dans la première phase
de calibrage, qui s'effectue hors ligne, les paramètres intrinsèques
de la caméra ainsi que la distorsion radiale sont estimés, et une
simple classification de couleurs est apprise à partir d'un exemplaire
d'une image de balles colorées. Ensuite, dans la phase de suivi en
ligne et temps réel, la classification des couleurs est appliquée
aux images en entrée, la balle est détectée et sa position 3D est
retournée. Dans la suite, nous expliquons en détails ces différentes
étapes.

\begin{figure}
\begin{centering}
\includegraphics[scale=0.4]{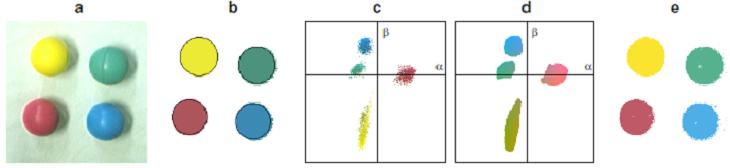}
\par\end{centering}

\caption{Calibrage couleur: (a) image couleur d'entrée, (b) segmentation d'image
par l'algorithme mean-shift et détermination des cercles, (c) distribution
non paramétrique de couleurs mesurée, (d) cinq groupes de couleurs
distincts, (e) classification pixel par pixel résultante.}
\end{figure}

\subsection{Calibrage}

La phase de calibrage hors ligne s'effectue en deux étapes. En premier
temps, les paramètres intrinsèques et la distorsion radiale de la
caméra sont estimés à partir de plusieurs images d'un damier prises
à partir de positions et d'orientations différentes. Nous exploitons
l'outil de calibrage de caméra GML {[}10{]} pour l'analyse et l'optimisation
de l'image. Les paramètres de la caméra extraits sont utilisés immédiatement
pour pré-calculer une table de correspondances qui permet de déterminer
les positions des pixels lors de l'exécution.

Dans la deuxième étape, nous estimons la distribution de couleur de
la balle qui doit être reconnue lors de la phase du suivi en ligne.
A cette fin, nous prenons une image couleur de la balle à partir de
la caméra (voir Figure 1(a)). Ensuite, nous effectuons une segmentation
couleur de l'image en utilisant une version modifiée de l'algorithme
mean-shift {[}11{]}. Dans notre cas, seule la consistance des couleurs
des régions est importante, raison pour laquelle l'algorithme des
mean-shift ne tient pas compte de la luminance et utilise seulement
les composantes couleurs de l'image (espace couleur LUV).
Cette modification, fait acquérir à l'algorithme de segmentation une
robustesse face aux variations de luminosité tout au long de la surface
de la balle (voir Figure 1b). 

Une fois l'image est segmentée, nous effectuons une analyse en composantes
connexes pour étiqueter les différentes régions. Dans
chaque région suffisamment large, un algorithme (voir section 3.2)
permet de déterminer si la région est un cercle ou non. Il permet
de générer un rapport d'ajustement qui est proche de 1 dans le cas
où la région est de forme circulaire. Dans ce cas, la région est reconnue
en tant que balle. Ensuite, les pixels détectés à l'intérieur du cercle
vont servir d'échantillons de couleur (voir Figure 1(c)) pour la distribution
non paramétrique représentée par une image $2D$, où les lignes et
les colonnes représentent les composantes de couleur $\alpha$ et
$\beta$ dans un espace de couleur sélectionné. On pourra utiliser
n'importe quel espace de couleur, il est envisageable par exemple
d'utiliser $a$ et $b$ de l'espace de couleur \emph{CIE Lab}. 

Quand tous les échantillons de couleur sont rassemblés, une opération
de fermeture est appliquée sur l'image pour remplir les petits trous
et filtrer le bruit (voir Figure 1d). Finalement, la distribution
non paramétrique de couleurs résultantes est utilisée pour calculer
une classification $RGB$ qui permet de convertir une couleur d'entrée
en indexe unique. La classification est implémentée comme étant une
table $3D$ de correspondance qui convertit les triplets $RGB$ en
des valeurs entières. Pour éviter une consommation abusive de la mémoire,
une représentation sur 6 bits est utilisée pour chaque composante
de couleur. La Figure 1(e) montre le résultat de la segmentation quand
la classification $RGB$ est appliquée sur l'image d'entrée.

\subsection{Suivi temps réel}

La phase de suivi temps réel consiste en quatre étapes principales;
(1) segmentation couleur dans laquelle l'image d'entrée est convertie
en plusieurs régions, (2) estimation robuste des paramètres du cercle,
(3) raffinement des paramètres du cercle, et (4) le suivi de la balle.
Dans ce qui suit, chaque étape est décrite avec plus de détails.

\begin{figure}
\begin{centering}
\includegraphics[scale=0.4]{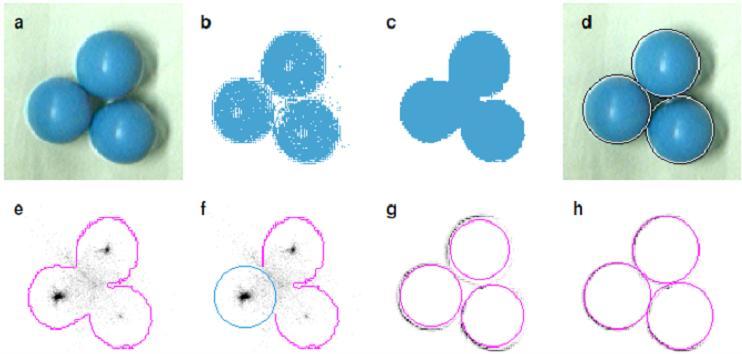}
\par\end{centering}

\caption{Détection de balle: (a) image couleur d'entrée, (b) classification
pixel par pixel, (c) réduction de bruit et remplissage de trous, (d)
détection des balles, (e) contours des régions et l'histogramme du
centre, (f) premier cercles détectés et élimination d'une partie du
contour d'origine, (g) Tous les cercles détectés et fermeture des
gradients de l'image, (h) contours ajustés aux contours réels.}
\end{figure}

\subsubsection{Segmentation}

Une fois l'image est acquise à partir de la caméra (voir Figure 2(a)),
on applique la classification RGB pour obtenir un seul indexe couleur
par pixel (voir Figure 2(b)). Ensuite, et comme déjà décrit dans la
phase de calibrage, une analyse en composantes connexes est effectuée
 pour obtenir une liste de régions et leurs relations de
voisinage. Après, on obtient un ensemble de régions dont l'une d'entre
elle au moins représente une balle (voir Figure 2(c)). Pour le reste
du traitement, chaque région est représentée par l'ensemble de ses
pixels de bord non déformés.

Le problème est maintenant plus simple si on le compare au problème
de recherche de cercle dans les images de contours lumineux. En effet,
nous allons effectuer une simple recherche dans un sous ensemble très
réduit de l'espace paramétrique 3D. Par contre, une estimation robuste
des paramètres du cercle s'avère nécessaire face à des problèmes comme
la présence d'autres objets de même couleur, l'occultation, où l'encombrement
de l'arrière plan.

Pour la détermination des paramètres du cercle, plusieurs approches
ont été utilisées. Citons, les approches qui se basent sur la transformée
de Hough, et celles basées sur les moindres carrés. Ces méthodes,
quoique robustes, leur convenance pour le temps réel reste discutable
dans quelques cas. Nous avons donc, essayé de développer une nouvelle
approche dont la première étape consiste à estimer de façon robuste
un cercle initial en utilisant un processus de vote aléatoire avec
réduction des dimensions de l'espace paramétrique. Ensuite les paramètres
du cercle sont raffinés en utilisant une technique par moindres carrés
pour un meilleur réajustement aux contours réels de la balle. Une
telle combinaison permet d'obtenir un équilibre entre robustesse,
précision et rapidité d'exécution. Chaque étape de l'algorithme proposé
va être décrite avec détails dans les paragraphes qui suivent.

\subsubsection{Estimation des paramètres du cercle}

Comme indiqué précédemment, l'hypothèse de base est que les limites
d'une des régions détectées trace exactement un cercle. Partant de
cette hypothèse, si nous prenons aléatoirement plusieurs pixels de
bord et nous traçons un cercle qui passe par ces pixels, la probabilité
d'avoir des cercles similaires répétitivement serait considérable.
Sachant que le calcul de la position du centre d'un cercle $(c_{x},c_{y})$
à partir de trois points $(x_{i},y_{i})$ avec des coordonnées entiers,
peut se faire rapidement :

{\large{
\begin{equation}
c_{x}=\frac{d_{1}y_{32}+d_{2}y_{13}+d_{3}y_{21}}{2(x_{1}y_{32}+x_{2}y_{13}+x_{3}y_{21})}
\end{equation}
}}{\large \par}

{\large{
\begin{equation}
c_{y}=\frac{d_{1}x_{32}+d_{2}x_{13}+d_{3}x_{21}}{2(y_{1}x_{32}+y_{2}x_{13}+y_{3}x_{21})}
\end{equation}
}}{\large \par}

Où $x_{ij}=x_{i}-x_{j}$ et $y_{ij}=y_{i}-y_{j}$ et $d_{i}=x_{i}^{2}+y_{i}^{2}$.
Par conséquence, nous pouvons prendre aléatoirement plusieurs votes
et recueillir les solutions dans un histogramme 2D avec la même résolution
que l'image originale (voir Figure 2(e)). Le processus de vote est
arrêté quand l'accumulateur de l'entrée incrémentée dépasse une limite
de votes ou quand on dépasse un total bien déterminé de votes. Ce
dernier cas est rare et survient lorsque les régions de deux ou plusieurs
balles interfèrent dans l'image (voir Figure 2(e,f)). Une fois le
cercle est estimé, nous procédons à un simple vote déterministe pour
calculer le rayon, nous calculons alors la distance (en pixels) de
chaque pixel de contour vers le centre du cercle et on incrémente
au même temps l'accumulateur de l'entrée correspondante dans un histogramme
1D. Finalement, on retient l'indexe de l'entrée qui a eu le plus de
votes comme une estimation du rayon du cercle.

Un problème qui peut survenir est l'existence d'un autre objet non
circulaire de même couleur que la balle dans l'image. Dans ce cas,
la phase de vote finira par dépasser un nombre maximum de votes aléatoires,
et l'histogramme de votes sera bruité sans aucun pic signifiant et
le maximum global sera considérablement faible (voir Figure 3). Pour
reconnaître cette situation on propose une métrique robuste qui témoigne
de la qualité du cercle :

{\large{
\begin{equation}
Q_{c}=\frac{c_{max}r_{max}}{N_{votes}N_{points}}
\end{equation}
}}{\large \par}

Ici $c_{max}$ et $r_{max}$ sont les maxima globaux des histogrammes
du centre et du rayon du cercle. $N_{votes}$ est le nombre de votes
du centre du cercle, et $N_{points}$ est le nombre des pixels de
contour de la région. Typiquement $Q_{c}$ pour les régions quasi
circulaires est quatre fois plus grande que celle pour les régions
non circulaires. La situation la plus douteuse est lorsque il y'a
interférence de la balle avec une région de même couleur, cependant
dans ce cas aussi l'amplitude de $Q_{c}$ est deux fois plus grande
(voir Figure 2(c)). 

\begin{figure}
\begin{centering}
\includegraphics[scale=0.5]{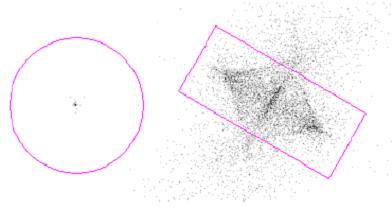}
\par\end{centering}

\caption{Histogramme de vote pour le centre d'un cercle: histogramme avec un
pic franc pour une région circulaire (gauche); l'histogramme est bruité
dans le cas où la région n'est pas circulaire et (droite)}
\end{figure}

\subsubsection{Raffinement des paramètres du cercle}

Une fois l'estimation du cercle est effectuée, nous procédons à un
raffinement de son centre et de son rayon pour les faire correspondre
au mieux au contour et au centre réels de la balle. Pour simplifier
cette tâche, nous assumons que les pixels de contour de la balle ont
un gradient d'intensité relativement élevé et sont au même temps près
du cercle estimé. Selon cette hypothèse de base, on applique la méthode
des différences centrées sur un petit anneau autour du cercle estimé
(voir Figure 2(g)) et on sélectionne les pixels avec les réponses
les plus grandes. Généralement, cette approche ne garantit pas qu'on
sélectionne toujours le contour réel du cercle puisque des problèmes
comme l'occultation ou l'encombrement du fond de l'image peuvent violer
notre hypothèse de base. Toutefois, la forme de l'anneau généralement
restreint la surface de sélection de façon à ce que les valeurs marginales
n\textquoteright{}affectent pas de façon sensible la solution finale.

Nous formulons la tâche comme étant une optimisation non linéaire
de moindres carrés avec la fonction d'énergie suivante:

{\large{}
\begin{equation}
E(c)=\underset{{\scriptstyle p}}{\sum}\left\Vert \nabla I.C(c)\right\Vert ^{2}\label{eq:E(c)}
\end{equation}
}{\large \par}

Où $\nabla I$ est le gradient de l'image au pixel $p$ et $C(c)$
est la distance du pixel $p$ au cercle $c$:

{\large{
\begin{equation}
C(c)=\sqrt{(p_{x}-c_{x})^{2}+(p_{y}-c_{y})^{2}}-c_{r}\label{eq:V distance}
\end{equation}
}}{\large \par}

Pour minimiser une telle fonction, nous utilisons la méthode itérative
de Gauss-Newton {[}12{]}. Où dans chaque itération les dérivés
du premier ordre de (\ref{eq:E(c)}) sont mis à zéro et la fonction
est linéarisée en utilisant un développement de Taylor du premier
ordre:

{\large{
\begin{equation}
\nabla E(c)=\underset{{\scriptstyle p}}{2\sum}\nabla C.\left\Vert \nabla I\right\Vert ^{2}.(C+\nabla C.\triangle c)=0
\end{equation}
}}{\large \par}

Avec:\foreignlanguage{english}{{\large{
\begin{equation}
\nabla C=\left(\frac{1}{d}\left(c_{x}-p_{x}\right),\frac{1}{d}\left(c_{y}-p_{y}\right),-1\right)\label{eq:Delta C}
\end{equation}
}}}{\large \par}

et

{\large{
\begin{equation}
d=\sqrt{(p_{x}-c_{x})^{2}+(p_{y}-c_{y})^{2}}
\end{equation}
}}{\large \par}

A partir de cette équation nous pourrons obtenir facilement le déplacement
incrémental des paramètres du cercle dans la notation matricielle:

{\large{
\begin{equation}
\triangle c=-\left(\nabla C^{t}.W.\nabla C\right)^{-1}\left(\nabla C^{t}.W.C\right)\label{eq:Calcul}
\end{equation}
}}{\large \par}

Où $\nabla C$ est une matrice $3\times N$ de dérivés du premier
ordre $3\times N$ (\ref{eq:Delta C}), $W=diag()\left\Vert \nabla I\right\Vert ^{2}$
est une matrice de pondération diagonale $N\times N$, et $C$ un
vecteur colonne $1\times N$ de distances des points au cercle $c$
(\ref{eq:V distance}).

En pratique, le calcul de (\ref{eq:Calcul}) est très rapide même
pour un nombre étendu de pixels, puisque une seule inversion de matrice
de taille $3\times3$ est calculée et une seule itération est nécessaire
pour réduire le décalage avec une précision sub-pixel (voir Figure
2(h)). Par ailleurs, nous avons remarqué dans nos expériences que
ces itérations peuvent être coûteuses dans le cas où des valeurs marginales
ou un fond encombré sont enregistrés dans l'image originale, c'est
pourquoi nous suggérons d'effectuer une seule itération même si le
coût en temps d'exécution n'est pas si élevé. 

Une fois, les paramètres du cercle $c_{x},$ $c_{y}$et $c_{z}$ sont
déterminés, la position 3D du centre peut être estimée en utilisant
un modèle de perspective:

{\large{
\begin{equation}
\begin{array}{ccc}
x=c_{x}\frac{z}{f}, & y=c_{y}\frac{z}{f}, & z=r\sqrt{1+\frac{f^{2}}{c_{r}^{2}}}\end{array}
\end{equation}
}}{\large \par}

Où $r$ est le rayon de la balle réelle et $f$ est la distance focale
de la caméra.

\subsubsection{Suivi de la balle}

Pour suivre les instances de la balle de manière cohérente et pour
éviter les échecs de détection à court terme causés par les occultations
de la balle ou par des changements brusques de luminosité, nous essayons
de faire fonctionner cette approche à une fréquence haute et nous
enregistrons à chaque fois la dernière position de la balle pour déterminer
la nouvelle position en cas de confusion avec un autre objet de même
classe de couleur et de même forme.

\section{Résultats expérimentaux}

Dans cette section nous essayons de décrire par des expériences la
performance et la précision de notre algorithme. Nous présentons aussi
les limites révélées par l'expérimentation. Enfin, nous essayons de
montrer les résultats donnés par cette approche sur le simulateur
du robot Pioneer 3.

\subsection{Performance et précision}

Le principal avantage de l'algorithme proposé par rapport aux approches
précédentes est le temps d'exécution global. En moyenne, cet algorithme
prend 2ms pour analyser une image de 0.7 Mpix avec une balle rouge
en utilisant un ordinateur double coeur (2.3GHz, 1GHz, FSB, 2MB Cache). 

La phase d'apprentissage hors ligne, qui renferme la classification
des pixels et l'analyse en composantes connexes, consomme en moyenne
6ms par image et peut être considérée comme un chargement constant.
L'estimation de la position de la balle a l'influence la plus importante
sur la vitesse de traitement. Elle prend en moyenne 0.5ms pour une
image dont la région de la balle est modérément grande et avec un
seuil de 16 pour l'accumulateur du centre du cercle. Pour atteindre
cette valeur, il faut en moyenne effectuer environ 300 votes aléatoires
pour le centre du cercle. 

La précision de l'algorithme dépend principalement de la résolution
de la caméra utilisée, des conditions d'éclairage et du degré d'occultation
de la balle. Dans nos expérimentations, nous avons utilisé une caméra
avec une résolution de 640x480. Grâce à la phase de raffinement des
paramètres du cercle, la déviation entre la balle estimée et la balle
réelle ne dépasse typiquement pas le 1mm. Cette valeur est valable
pour une balle non occultée de 7cm de diamètre éclairé avec une lumière
du jour, positionnée à une distance de 1m de la caméra (diamètre de
60 pixels dans l'espace de l'image). Dans le cas d'une occultation
importante (plus de 50\%) et/ou de mauvaises conditions d'éclairage,
la déviation par rapport à la position réelle peut être encore plus
grande.

\subsection{Limites de l'approche}

Mise à part la non généricité de l'approche qui ne détecte que les
formes circulaires, la principale limitation de l'algorithme proposé
est qu'il repose sur l'hypothèse qui stipule que la lumière entrante
ainsi que le spectre de couleurs sont constants. Cette limitation
est étroitement liée au nombre de couleurs distinctes qui devraient
être reconnues. Il y'a un compromis entre le nombre de couleurs actives
et la taille des classes de couleurs. Une grande taille de la classe
apporte plus de robustesse aux changements de couleur mais diminue
le nombre de couleurs distinctes et augmente la probabilité de collision
avec des objets de fond. D'après nos expériences 4 classes de couleurs
distinctes permettent un bon compromis entre le nombre de couleurs
et la robustesse du système. Si la lumière du jour est utilisée pour
l'éclairage et plusieurs couleurs sont exigées pour l'apprentissage,
il faut exécuter le calibrage à plusieurs reprises dans différents
moments de la journée. 

Une autre limitation de cette approche est la précision des paramètres
estimés de la balle. Quand la balle est très loin de la caméra ou
fortement occultée, l'estimation de la position du cercle et du rayon
peut être erronée et la profondeur qui en résulte peut être notablement
différente de la profondeur réelle. Une situation similaire se produit
également, lorsque des mauvaises conditions d'éclairage et/ou des
changements dans le spectre de la lumière génèrent une classification
bruitée de couleurs.

\subsection{Simulation du suivi temps réel}

Nous avons utilisé un simulateur  du robot mobile Pioneer 3. Une
webcam est fixé au dessus du Pioneer 3 et une balle rouge est utilisée
comme cible pour le suivi. Nous utilisons une seule balle, donc seulement
deux classes de couleurs sont à définir; celle des pixels de la balle
et celle des pixels qui n'appartiennent pas à la balle.

La Figure 4 montre les étapes de traitement de l'image pour la détection
de la balle. La figure 4(a) montre l'image de la balle prise par la
caméra avec des objets de couleur semblable à la balle. L'image 4(e)
montre la transformation de la balle en image en noir et blanc selon
l'apprentissage de la couleur de la balle (l'étape hors ligne). Les
images 4(c) et 4(d) sont obtenues après ouverture, fermeture et élimination
du bruit sur l'image 4(b). L'image 4(e) montre la première estimation
des paramètres du cercle de la balle. L'image 4(f) montre le cercle
de l'image après raffinement des paramètres par la méthode déjà présentée.

\begin{figure}
\begin{centering}
\includegraphics[scale=0.27]{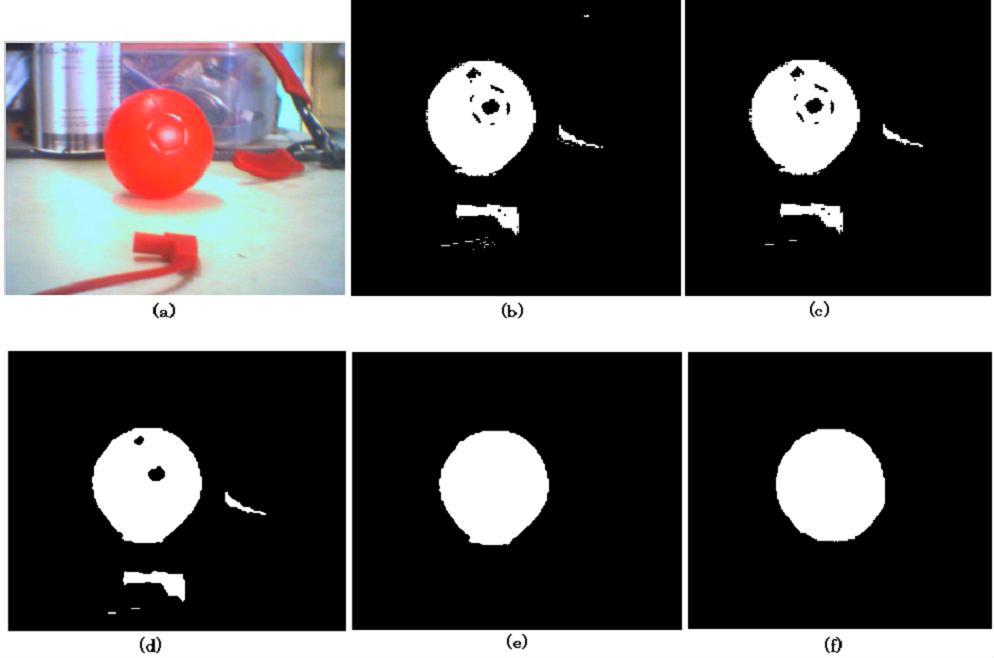}
\par\end{centering}

\caption{Détection de la balle : (a) image source; (b) image après
Segmentation; (c) image après ouverture;
(d) image après fermeture; (e) estimation des paramètres du cercle;
(f) raffinement du paramètres du cercle.}
\end{figure}

Le robot Pioneer 3 est équipé d'un module d'évitement d'obstacles
en temps réel en utilisant ses capteurs ultrasons. L'intégration du
module de suivi de la balle n'a pas affecté l'aspect temps réel et
le robot parvient à éviter les obstacles tout en suivant la balle
comme le montre la Figure 5.

\begin{figure}
\begin{centering}
\includegraphics[scale=0.3]{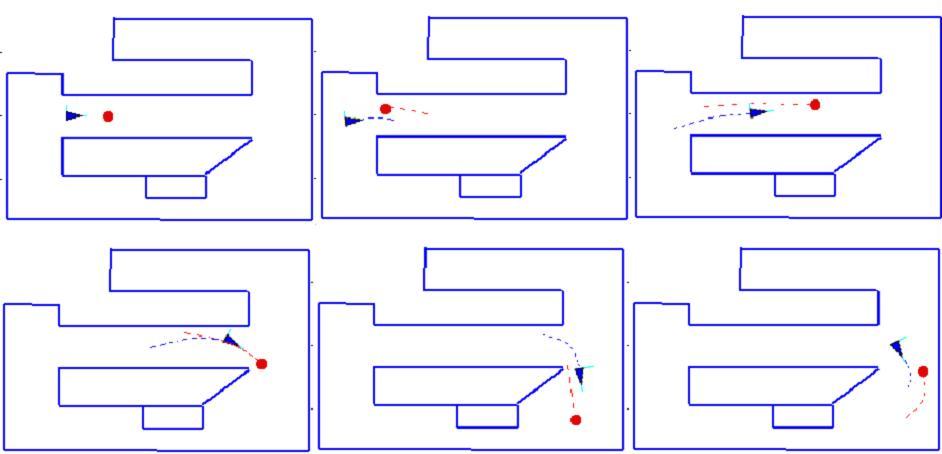}
\par\end{centering}

\caption{Suivi de la balle}
\end{figure}

\section{Conclusion}

Dans ce papier, nous avons présenté une classification des méthodes
de suivi d'objets selon les techniques et les approches utilisées.
Ensuite, nous avons proposée une approche hybride pour le suivi d'une
balle colorée en temps réel. Cette approche repose sur deux phases;
une phase d'apprentissage hors ligne qui permet de donner une classification
de couleurs selon les couleurs des cibles; et une phase en ligne qui
permet de détecter une ou plusieurs balles et donner avec précision
l'estimation des cercles correspondant. Les expériences effectuées
ont montré que cette approche convient d'être embarquée avec d'autres modules temps
réel sans autant affecter le temps d'exécution.


\begin{thebibliography}{9}
\bibitem{foo:baz}
D. M. Gámez, Michel Devy, ''Active visual-based detection
and tracking of moving objects from clustering and classification
methods'', \emph{Advanced Concepts for Intelligent Vision Systems},
Lecture Notes in Computer Science Volume 7517, pp 361-373, 2012.

\bibitem{key:foo}P. F. McLauchlan, et J. Malik, \textquotedblleft{}Vision
for Longitudinal Vehicle Control\textquotedblright{}, \emph{in Proceedings
of the Eighth British Machine Vision Conference (BMVC\textquoteright{}97)},
1997.

\bibitem{key-6}X. Li, W. Hu, C. Shen, Z. Zhang, A. R. Dick et A. Hengel, \textquotedblleft{}A Survey of Appearance Models in Visual Object Tracking{}, \emph{ ACM Transactions on Intelligent Systems and Technology}, Vol. 4 Issue 4, September 2013.


\bibitem{key-8}D. J. Koller, Weber, et J. Malik, \textquotedblleft{}Robust
multiple car tracking with occlusion reasoning\textquotedblright{}
\emph{in Proc. 3rd European Conf. on Computer Vision (ECCV\textquoteright{}94)},
Stockholm, vol. 1, pp. 189-196, May, 1994. 

\bibitem{key-9}S. M. Smith et J.M. Brady, \textquotedblleft{}ASSET-2:
Real-Time Motion Segmentation and Shape Tracking\textquotedblright{},
\emph{in IEEE. Trans. on Pattern Analysis and Machine Intelligence},
vol. 17, No. 8, pp. 814-820, August 1995.

\bibitem{key-10}B. Bascle, P. Bouthemy, R. Deriche, et F. Meyer,
\textquotedblleft{}Tracking complex primitives in an image sequence\textquotedblright{},
\emph{Technical Report 2428, INRIA, Sophia-Antipolis}, France, December
1994.

\bibitem{key-11}T. Drummond, et R. Cipolla, \textquotedblleft{}Real-time
tracking of complex structures with online camera calibration\textquotedblright{},
in Image and Vision Computing, vol. 20, No. 5-6, pp. 427-433, 2002.

\bibitem{key-12}F. Dellaert, C. Thorpe, and S. Thrun, \textquotedblleft{}Super-Resolved
Texture Tracking of Planar Surface Patches\textquotedblright{}, in
Proc. of IEEE/RSJ International Conference on Intelligent Robotic
Systems, October, 1998. 

\bibitem{key-13}N. X. Dao, BJ.You, SR.Oh and M. Hwangbo, \textquotedblleft{}Visual
Self-Localization for Indoor Mobile Robots Using Natural Lines\textquotedblright{},
\emph{in Proc. of the 2003 IEEE/RSJ Intl. Conference on Intelligent
Robots and Systems IROS\textquoteright{}2003}, pp. 1252-1257, Las
Vegas, Nevada, October, 2003.


\bibitem{key-15}V. Vezhnevets, A. Velizhev , ''GML C++ camera calibration
toolbox'', 2005.

\bibitem{key-16}D. Comaniciu, P. Meer: Robust analysis of feature
spaces, ''Color image segmentation''. \emph{In Proceedings of Conference
on Computer Vision and Pattern Recognition}, pp. 750\textendash{}755,
(1997). 

\bibitem{key-17}W. Press, S. Teukolsky, W. Vetterling, B. Flannery,
''Numerical Recipes in C: The art of scientific computing''. \emph{Cambridge
University Press}, 1992.

\bibitem{key-18}H. Ghazouani, "Navigation visuelle de robots
mobiles dans un environnement d'int\'erieur", Thèse de doctorat,
Universit\'e Montpellier II-Sciences et Techniques du Languedoc, 2012.

\bibitem{key-19}H. Ghazouani "Genetic Stereo Matching Algorithm with Fuzzy Fitness",
\emph{5th International Conference on Metaheuristics and Nature Inspired Computing (META'1 4)}, pp. 1-5, Tunis, Tunisia, 2014.

\bibitem{key-20}H. Ghazouani "Fast and
Robust Semi-Local Stereo Matching Using Possibility distributions",
\emph{International Journal of Computational Vision and Robotics (IJCVR)}
- Vol. 2, No.3 pp. 237-253, 2011.

\end{thebibliography}
\end{document}